\documentclass[a4paper, 10pt, conference]{ieeeconf}      
\usepackage{fancyhdr}
\usepackage{FG2024}
\usepackage{graphicx}
\usepackage{pifont}
\usepackage{amsmath}
\usepackage{amssymb}
\usepackage[font=footnotesize]{caption} 
\usepackage{cuted}
\usepackage{comment}
\usepackage{url}

\usepackage{hyperref}
\hypersetup{colorlinks=true,
citecolor=green,
urlcolor=magenta,
linkcolor=black,}

\makeatletter

\IEEEoverridecommandlockouts                              

\def\FGPaperID{224} 
\FGfinalcopy

\title{\LARGE \bf
In My Perspective, In My Hands: \\Accurate Egocentric 2D Hand Pose and Action Recognition
}

\author{\parbox{16cm}{\centering
   {\large Wiktor Mucha and Martin Kampel}\\
   {\normalsize
   Computer Vision Lab, TU Wien, Favoritenstr. 9/193-1, 1040 Vienna, Austria}}
}

\begin{document}


\maketitle

\begin{strip}\vbox{
\vspace{-1cm}

\includegraphics[width=\textwidth]{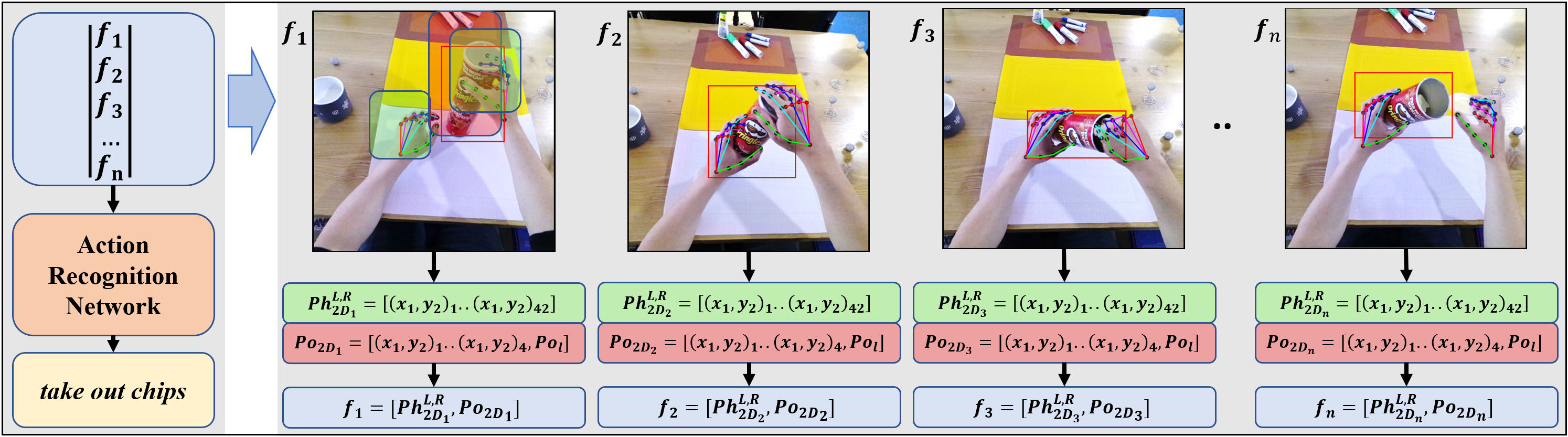}

\vspace{0.1cm}
\refstepcounter{figure}\footnotesize{Fig.~\thefigure. Overview of our method. From the sequence of input frames $f_1, f_2, f_3 .. f_n$ representing action, 2D hands pose $Ph_{2D}^L$,$Ph_{2D}^R$ and bounding box of manipulated objects $Po_{2D}$ with its labels $Po_{l}$ are extracted. Under the study, four distinct state-of-the-art hand pose methods are implemented and tested. Object information is retrieved using \textit{YOLOv7} \cite{wang2022yolov7}. Pose information is embedded into a vector describing each frame. The sequence of vectors is processed via the transformer-based deep learning neural network to predict the final action class.}
  \label{fig-teaser}

}
\end{strip}

\begin{abstract}

Action recognition is essential for egocentric video understanding, allowing automatic and continuous monitoring of Activities of Daily Living (ADLs) without user effort. Existing literature focuses on 3D hand pose input, which requires computationally intensive depth estimation networks or wearing an uncomfortable depth sensor. In contrast,  there has been insufficient research in understanding 2D hand pose for egocentric action recognition, despite the availability of user-friendly smart glasses in the market capable of capturing a single RGB image. Our study aims to fill this research gap by exploring the field of 2D hand pose estimation for egocentric action recognition, making two contributions. Firstly, we introduce two novel approaches for 2D hand pose estimation, namely \textit{EffHandNet} for single-hand estimation and \textit{EffHandEgoNet}, tailored for an egocentric perspective, capturing interactions between hands and objects. Both methods outperform state-of-the-art models on \textit{H2O} and \textit{FPHA} public benchmarks. Secondly, we present a robust action recognition architecture from 2D hand and object poses. This method incorporates \textit{EffHandEgoNet}, and a transformer-based action recognition method. Evaluated on \textit{H2O} and \textit{FPHA} datasets, our architecture has a faster inference time and achieves an accuracy of 91.32\% and 94.43\%, respectively, surpassing state of the art, including 3D-based methods. Our work demonstrates that using 2D skeletal data is a robust approach for egocentric action understanding. Extensive evaluation and ablation studies show the impact of the hand pose estimation approach, and how each input affects the overall performance. The code is available at \small{\url{https://github.com/wiktormucha/effhandegonet}}.

\end{abstract}

\section{Introduction}


The growing interest in egocentric vision research is evident from the release of large dedicated datasets like \textit{EPIC-KITCHENS}~\cite{Damen2018EPICKITCHENS}, \textit{Ego4D}~\cite{grauman2022ego4d}, and \textit{H2O}~\cite{Kwon_2021_ICCV}. One of the challenges in this domain is the task of action recognition, focused on determining the action performed by the user in the video~\cite{nunez2022egocentric}.
Research in egocentric action recognition is crucial due to broad potential application fields, including augmented and virtual reality, nutritional behaviour analysis, and Active Assisted Living (AAL) technologies for lifestyle analysis~\cite{nunez2022egocentric} or assistance~\cite{mucha2024text2taste}. ADLs targeted by AAL technologies (e.g., drinking, eating and food preparation) are all based on manual operations and manipulations of objects, which motivates research focused on hand-based action recognition.

Current works on egocentric action recognition focus on 3D hand pose~\cite{tekin2019h+,das2021symmetric,Kwon_2021_ICCV}, despite the absence of wearable depth sensors in the market. Consequently, these studies resort to estimating depth from RGB frames, introducing complexities and yielding pose prediction errors around 40 mm~\cite{tekin2019h+,Kwon_2021_ICCV} (equivalent to a 20.5\% error, considering an average human hand size of 18 cm).  While depth maps could be directly acquired through sensors, this necessitates inconvenient custom setups, as illustrated in Fig. \ref{wearabledepthsensor}. As an alternative to 3D hand pose estimation, 2D estimations are reported to be more accurate at the time of this study considering percental error (13.4\% for 2D~\cite{zhang2020mediapipe} against 20.5\% for 3D~\cite{tekin2019h+,das2021symmetric,Kwon_2021_ICCV}).
Additionally, the existing literature provides examples of 2D-based action recognition achieving higher accuracy in non-egocentric setups compared to 3D-based methods~\cite{duan2022revisiting}.  These factors underscore the need for further exploration of the potential advantages of 2D pose estimation for egocentric action recognition.

Our study, based on 2D keypoints, investigates this gap with the goal of bridging the distance between research and practical applications, enabling the utilization of off-the-shelf RGB egocentric cameras. An overview of our approach is shown in Fig. \ref{fig-teaser}, which displays the 2D hand and object poses obtained by our methods. It includes an example of a sequence representing an action captured from an egocentric perspective of the \textit{H2O Dataset}~\cite{Kwon_2021_ICCV}.  The proposed method is applicable to recordings from recently released wearable RGB smart-glasses like RayBan Stories\footnote{\label{rayban} RayBan Stories - \url{https://www.ray-ban.com/usa/ray-ban-stories} (accessed 03 July 2023)} and Snapchat Spectacles\footnote{Snapchat Spectacles - \url{https://www.spectacles.com} (accessed 03 July 2023)}. These devices allow the use of only a single RGB image, despite incorporating dual cameras, and offer improved comfort, quality and lightweight design for egocentric vision compared to head-mounted cameras or self-made RGB-D sensors (see Fig. \ref{wearabledepthsensor}). This is particularly noteworthy given the lack of commercially available wearable RGB-D devices on the market at the time of the study. The release of these user-friendly glasses is expected to increase the availability of single-image RGB datasets, thus fostering egocentric vision research. Lastly, works on 3D lifting~\cite{mehraban2024motionagformer}, which are the current state of the art for 3D human pose estimation, highlight the importance of accurate 2D pose predictions as input for the generation of 3D poses using lifting algorithms. Our 2D hand pose estimation method is proven to predict accurate 2D poses, which means that our method is not limited to direct action recognition from 2D keypoints, but holds potential as a versatile first step for 3D pose estimation through lifting algorithms.

 Our contributions are the following:

\begin{figure}[t]
  \centering
  \includegraphics[width=\linewidth]{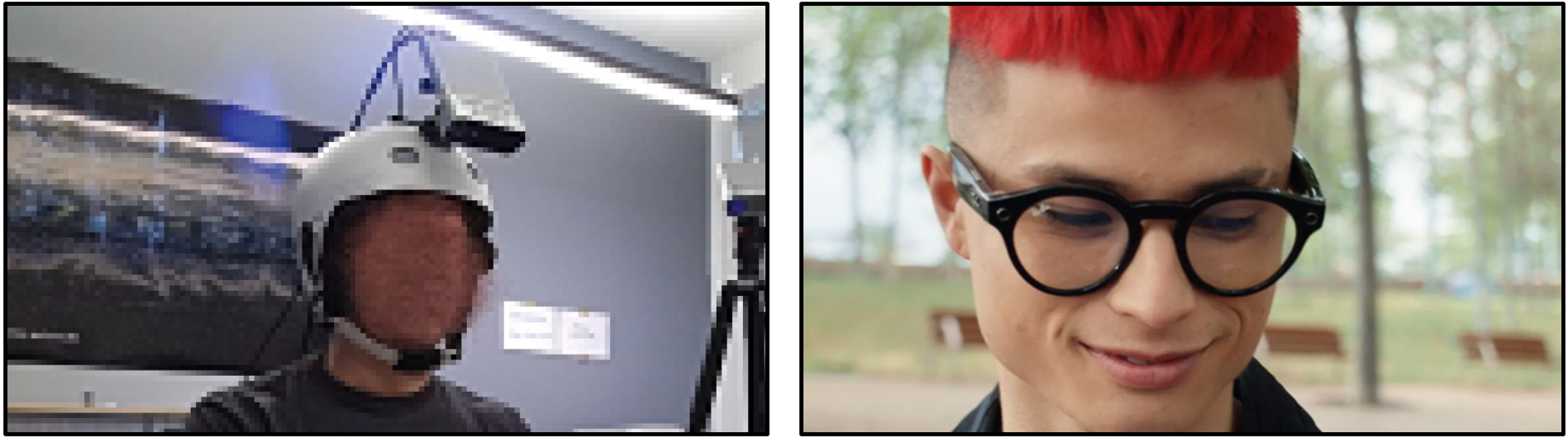}
  \caption[Caption for LOF]{Currently, comfortable wearable RGB-D cameras are not readily available in the market. Left: A self-made RGB-D setup~\cite{Kwon_2021_ICCV}. Right: User wearing RayBan glasses with an integrated RGB camera\footref{rayban}. }
  \label{wearabledepthsensor}
  \vspace{-0.25cm}
\end{figure}

\begin{enumerate}
  
\item A state-of-the-art architecture for single 2D hand pose prediction, \textit{EffHandNet}, surpassing other methods in every metric on the single-hand \textit{FreiHAND}~\cite{zimmermann2019freihand}. 

\item A novel architecture for modelling 2D hand pose from the egocentric perspective, \textit{EffHandEgoNet} that outperforms state-of-the-art methods on \textit{H2O}~\cite{Kwon_2021_ICCV} and \textit{FPHA}~\cite{garcia2018first} datasets. This includes models for both 2D and 3D pose estimation, with the latter projected into 2D space for a fair comparison.

\item A novel method for egocentric action recognition using 2D data on the subject's hands and object poses from \textit{YOLOv7}~\cite{wang2022yolov7}. 
Our method distinguishes itself from other studies by employing a reduced set of input information, leading to faster inference per action. By leveraging \textit{YOLOv7}, which has demonstrated excellent performance at recognizing diverse labels in large datasets, we enhance the potential for easy generalization to various tasks and datasets in the future. A comprehensive evaluation on the \textit{H2O Dataset} showcases superior accuracy, reaching 91.32\%, outperforming other state-of-the-art methods. In single-handed experiments on the \textit{FPHA Dataset}, our method demonstrates robustness, achieving a state-of-the-art performance of 94.43\% when only one hand manipulates an object.

\item We present extensive experiments and ablations performed on \textit{H2O Dataset}, showing the influence of the hand pose method by comparing four different pose estimation techniques. Further analysis shows the importance of each input (left and right hand, object position) for the results of action recognition.

\end{enumerate}

The paper is organised as follows: Section \ref{related-work} presents related work in egocentric hand keypoints estimation and hand-based action recognition highlighting areas for improvement. Section \ref{methodology} describes our methods and implementation. Evaluation and experiments are presented in Section \ref{evaluation}. Section \ref{conclusion} concludes the study summarising its main findings.


\section{Related Work}
\label{related-work}

Action recognition has been extensively studied in the literature, with approaches employing diverse modalities such as RGB images, depth information, and skeletal data \cite{sun2022human}. In our study, we focus on egocentric hand-based action recognition using 2D keypoints estimated from a single RGB image. Therefore, we present related work in egocentric vision for hand pose estimation and hand-based action recognition. 

\subsection{Egocentric Hand Keypoint Description}

Hand-pose estimation in egocentric vision is challenged by self-occlusion during movements, limited field of view, and diverse perspectives, making effective generalisation difficult. Some works address these challenges by employing RGB-D sensors~\cite{mueller2017real,garcia2018first}. In addition, the depth modality could enhance user privacy~\cite{mucha2022addressing, mucha2022beyond}, but the adoption of depth-sensing devices in the market is limited, requiring users to wear uncomfortable devices (see Fig. \ref{wearabledepthsensor}). 

To leverage the benefits of 3D keypoints, some researchers employ neural networks for depth estimation and subsequent conversion from 2D to 3D space using intrinsic camera parameters~\cite{tekin2019h+,Kwon_2021_ICCV}. For instance, Tekin~et~al.~\cite{tekin2019h+} use a single RGB image to directly calculate the 3D pose of a hand through a Convolutional Neural Network (CNN) which generates a 3D grid as the output of CNN, where each cell contains the probability of target pose value. Kwon~et~al.~\cite{Kwon_2021_ICCV} follow this approach but estimates pose for both hands. This work, however, reports a mean End-Point Error (EPE) of 37mm for hand pose estimation in the \textit{H2O Dataset}, which, considering the average human hand size of 18 cm, results in a 20\% error, leaving room for improvement. Cho~et~al.~\cite{cho2023transformer} uses CNN with the transformer-based network for 3D pose-per-frame reconstruction. Wen~et~al.~\cite{wen2023hierarchical} propose to use sequence information to reconstruct depth omitting the issue of occlusions. However, these works focus on 3D pose estimation from the egocentric perspective. Despite the practicality of RGB camera glasses for real-world applications, 2D hand-based approaches are not widely adopted in the egocentric domain.  One of a few works in egocentric 2D hand pose is an early study by Liang~et~al.~\cite{liang2015egocentric}. A single RGB camera is employed to estimate hand pose and its distance from the sensor, employing a Conditional Regression Forest; however, this study lacks quantitative results. Another approach by Wang~et~al.~\cite{wang2018mask} introduces a cascaded CNN for 2D hand pose prediction, involving two stages: hand mask creation and hand pose prediction.

Beyond the egocentric domain, a review of available applications reveals that improvements in network architectures for regular pose estimation and keypoint prediction are also applicable to the egocentric hand pose prediction. This is demonstrated in the study by Bauligu~et~al.~\cite{baulig2018adapting}, who successfully adapt OpenPose~\cite{cao2017realtime} to the egocentric perspective. However, despite continuous improvements in single-hand pose performance, its application in the egocentric environment remains relatively unexplored, prompting the focus of this study. Recent advancements in 2D hand pose estimation include \textit{PoseResNet50}~\cite{mmpose2020}, which is based on residual connection architecture and Santavas~et~al.~\cite{santavas2020attention}, who reinforce a CNN by a self-attention module. Zhang~et~al.~\cite{zhang2020mediapipe} proposes \textit{MediaPipe}, which employs single shot detection to identify hand regions passed further through the network to estimate the final pose for each hand.

Our contribution differs from available works, focusing on 2D prediction based on a single RGB image. We implement top-down and bottom-up methods and compare them with state-of-the-art in the egocentric environment presenting quantitative and qualitative results [supplement]. We assess their transition from a standard to an egocentric perspective by employing them to build a 2D hand-based action recognition system.

\subsection{Hand-Based Egocentric Action Recognition}

A common strategy for action recognition involves processing jointly hand and object information. Cartas~et~al.~\cite{cartas2017contextually} propose CNN-based object detectors to estimate the positions of primary regions (hands) and secondary regions (objects). Temporal information from these regions is then processed by a Long Short-Term Memory (LSTM) network. Nguyen~et~al.~\cite{nguyen2019neural} transition from bounding box information to 2D skeletons of a single hand estimated by CNN from RGB and depth images. The joints of these skeletons are aggregated using spatial and temporal Gaussian aggregation, and action recognition is performed using a learnable Symmetric Positive Definite (SPD) matrix.

With the rise of 3D-based hand pose estimation algorithms, the scientific community has increasingly focused on egocentric action understanding using 3D information~\cite{tekin2019h+,das2021symmetric,Kwon_2021_ICCV}. Tekin~et~al.~\cite{tekin2019h+} estimate 3D hand and object poses from a single RGB frame using CNN, embedding temporal information for predicting action classes using LSTM. Other techniques employ graph networks, such as Das~et~al.~\cite{das2021symmetric}, who present an architecture with a spatio-temporal graph CNN, describing finger movement using separate sub-graphs. Kwon~et~al.~\cite{Kwon_2021_ICCV} construct sub-graphs for each hand and object, which are merged into a multigraph model, allowing for learning interactions between these components. Wen~et~al.~\cite{wen2023hierarchical} use a transformer-based model with estimated 3D hand pose and object label input. Cho~et~al.~\cite{cho2023transformer} enrich the transformer inputs with object pose and hand-object contact information. However, these studies do not employ sensor-acquired depth data. Instead, they estimate points in 3D space using neural networks and intrinsic camera parameters~\cite{tekin2019h+,Kwon_2021_ICCV,wen2023hierarchical,cho2023transformer}.

In contrast, our study aims to fill the gap of egocentric action recognition based on 2D hand pose. We introduce a novel architecture based on 2D keypoints, estimated with a lower percentage error in pose prediction compared to 3D-based methods that outperform existing solutions. It effectively leverages information from both hands and relies solely on RGB images. Moreover, by bypassing the architecture responsible for lifting points to the third dimension, we reduce network complexity and achieve a faster inference time per action, which is a critical factor in certain applications.

\section{Egocentric Action Recognition Based on 2D Hand Pose}
\label{methodology}

\label{sec_model_arch}

\begin{figure*}[t]
  \centering
  \includegraphics[width=\linewidth]{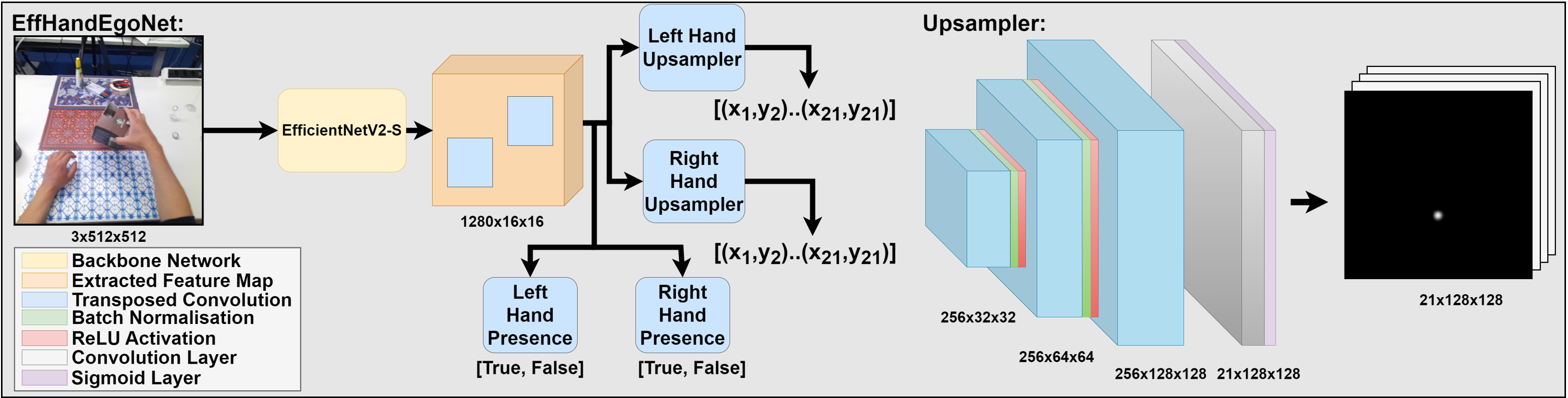}
  \caption{Our \textit{EffHandEgoNet} architecture to resolve keypoint prediction problem. Input images are resized to 512 pixels and passed through the network to produce 21 heatmaps for each of the hand's keypoints by both upsamplers. The size of the layers is illustrative.}
  \label{fig:effhandegonet}
  \vspace{-0.25cm}
\end{figure*}

The proposed method performs action recognition from image sequences by processing pose information describing hands and interacting objects in a 2D space. We consider fine-grained actions of a user manipulating different objects with his hands, e.g. opening a bottle or pouring milk. The pipeline overview is presented in Fig. \ref{fig-teaser} where $\mathit f_{n}$ corresponds to processed frames. It is constructed of three separate blocks object detection, hand pose estimation and finally, action recognition using transformer encoder block and fully connected layers.

\subsection{Object Detection and Pose Estimation}

The first step in the pipeline is object detection, which is carried out employing the pre-trained \textit{YOLOv7} network \cite{wang2022yolov7}. Following that, transfer learning is used on the \textit{H2O Dataset}, using the Ground Truth (GT) object pose. This process involves transforming from a 6D GT representation of objects to a 2D bounding box.
In each frame, denoted as $\mathit f_n$, the interacting object is represented by $\mathit{Po}_{2D}(x,y) \in \mathbb{R}^{4 \times 2}$, where each point corresponds to the corners of its bounding box. Additionally, $\mathit{Po}_{l} \in \mathbb{R}^{1}$ represents object's label.

\subsection{Hand Pose Estimation}

Each frame $\mathit f_n$ contains alongside the object pose, the hands pose of a subject conducting action.
This hand pose is estimated with an independent network part dedicated to inferring the position of 21 keypoints $j$ in each hand.
Each point $j$ expresses the position of wrist and fingers' joints following the standard approach of hand pose description. 
Each hand pose is represented by $\mathit{Ph}_{2D}^{L,R}(x,y), \in \mathbb{R}^{j \times 2}$ where 
$L$ and $R$ describes the left or right hand and $x, y$ represents the point $j$.

\subsubsection{EffHandNet - Top-Down Single Hand Model}

Hand instances appearing in the scene are detected and extracted from the image as $R_{L}, R_{R} \in \mathbb{R}^{3\times w\times h}, w,h = 128$ using finetuned \textit{YOLOv7} \cite{wang2022yolov7}. Further, these regions $R_{L}, R_{R}$ are passed to the model that predicts hand keypoints in a single-hand image. Image features $F_M \in \mathbb{R}^{1280\times 4 \times 4}$ are extracted with \textit{EfficientNetV2-S} \cite{tan2021efficientnetv2}. It provides state-of-the-art accuracy with a medium amount of trainable parameters (21.45M parameters). Further, the feature matrix $F_M$  is passed via an upsampler which generates heatmap $\mathit H\in \mathbb{R}^{J\times w\times h}$ where each cell represents the probability of joint $J$ occurrence. We create \textit{EffHandNet} model with the upsampler containing a stack of five transposed convolutions and pointwise convolution for channel reduction and probability layer. Finally, $H$ is transformed into $\mathit P_{2D}$ choosing highest probability cell for each $J$.

\subsubsection{EffHandEgoNet - Bottom-Up Egocentric Model}

The challenge to overcome in the egocentric view concerning hand pose description is the accurate modelling of interactions between two hands and objects, where the top-down approach tends to fail. To close this performance gap, we introduce \textit{EffHandEgoNet}, a modified \textit{EffHandNet} to execute bottom-up keypoints estimation in the egocentric perspective. It simultaneously allows the processing of two interacting hands in the scene, improving the pose prediction robustness. The network consists of \textit{EfficientNetV2-S} backbone extracting features from the image $I \in \mathbb{R}^{3\times w\times h}, w,h = 512$. Extracted features $F_M \in \mathbb{R}^{1280\times 16 \times 16}$ are handed to two independent upsamplers for each of the hands, following the accurate \textit{EffHandNet} approach and to the handness modules responsible for predicting each hand's presence $h^L, h^R \in \mathbb{R}^2$ built from linear layers. The upsamplers consist of three transposed convolutions with batch normalisation and ReLU activation except the last layer followed by a pointwise convolution.  
Heatmaps $H^L$ and $H^R$ are transformed into $\mathit P_{2D}^{L}, P_{2D}^{R}$.
The full architecture is illustrated in Fig. \ref{fig:effhandegonet}.

\subsection{Action Recognition}

\begin{figure*}[t]
\begin{center}
\includegraphics[width=\textwidth]{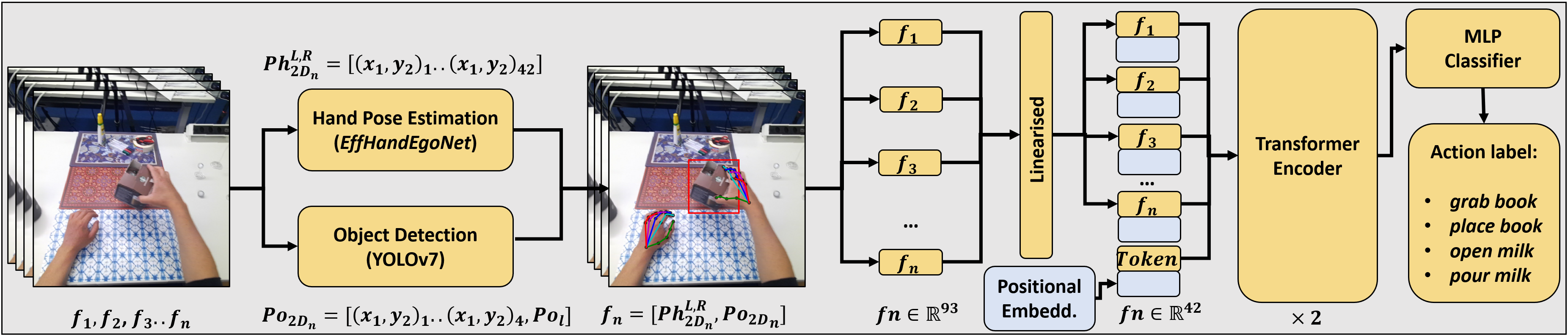}
\vspace{-0.6cm}
\end{center}
 \caption{ Our procedure for action recognition. From the sequence of frames $f_1, f_2, f_3 .. f_n$ hand pose $Ph_{2D}^L$,$Ph_{2D}^R$ is estimated using \textit{EffHandEgoNet} model and object pose $Po_{2D}$, $Po_{l}$ is extracted with \textit{YOLOv7} \cite{wang2022yolov7}. Each sequence frame $ f_n $ is linearised from shape $\mathbb{R}^{93}$ to $\mathbb{R}^{42}$. With added positional embedding and classification token, this information creates an input for the transformer encoder implemented following \cite{dosovitskiy2020image} repeated $\times2$ times, which embeds the temporal information. Finally, the multi-layer perceptron predicts one of the 36 action labels.}
\label{action_diagram}
\vspace{-0.25cm}
\end{figure*}

The representation of each action sequence consists of frames $\mathit [f_1,f_2,f_3..f_n]$, where $n \in [1..N]$ and $N=20$ is chosen heuristically. These frames embed flattened poses of hands $Ph_{2D}^{L}$, $Ph_{2D}^R$ and object $Po_{2D}$,  $Po_{l}$. If fewer than $N$ frames represent an action, zero padding is applied, while actions longer than $N$ frames are sub-sampled.
The input vector $\mathit V_{seq}$ is a concatenation of frames $\mathit f_n \in \mathbb{R}^{93} $.
\begin{equation}
f_n =[ Ph_{2D}^{L}, Ph_{2D}^{R}, Po_{2D}, Po_{l}]
\end{equation}
\begin{equation}
V_{seq} = [f_1, f_2..f_n], n \in [1..20]
\end{equation}

The final stage of the pipeline involves processing the sequence vector $\mathit V_{seq}$ to embed temporal information and perform action classification. For this objective, we employ a model inspired by the Visual Transformer \cite{dosovitskiy2020image}. It is constructed using a standard Transformer Encoder block \cite{vaswani2017attention}.
The input vector $\mathit V_{seq}$ representing an action is linearised using a fully connected layer to $\mathit x_{lin}$. The resulting $\mathit x_{lin}$ is combined with a classification token, and a positional embedding following \cite{dosovitskiy2020image}.
The number of encoder layers is set to 2, and encoder dropout and attention dropout are applied with a probability of 0.2. The number of network parameters is equal to only 31K and the inference time on the RTX 3090 GPU equals 6.2 ms. The detailed pipeline is depicted in Fig. \ref{action_diagram}.

\section{Experiments}
\label{evaluation}


The proposed action recognition approach is evaluated on two different datasets. \textit{H2O Dataset} \cite{Kwon_2021_ICCV} contains actions performed with both hands and \textit{FPHA Dataset} \cite{garcia2018first} contains actions performed with only one hand. The choice of these datasets is motivated by following other studies in the field of hand-based egocentric action recognition \cite{Kwon_2021_ICCV,wen2023hierarchical} to allow comparison with existing work. Besides this, the provided GT hand pose and object information allows us to perform extensive experiments and ablations regarding different inputs which are not possible in other egocentric datasets, i.e., \cite{Damen2018EPICKITCHENS, grauman2022ego4d}. In addition, single-hand estimation task is evaluated on the non-egocentric \textit{FreiHAND} \cite{zimmermann2019freihand}.

\subsection{Datasets}

\subsubsection{\textbf{FreiHAND}} It is a dataset for hand pose and shape estimation from a single RGB image. In this study, it is used for the training and evaluation of a network responsible for detecting single-hand keypoints. It contains 130K images of hands annotated with 3D coordinates that can be transformed to 2D using the given camera intrinsics. There are 32K images of hands captured on a green screen, and the rest are created by background augmentation. In addition, there are 4K images for testing, including real backgrounds referenced in this study as \textit{final test}. 

\subsubsection{\textit{\textbf{H2O Dataset}}} It provides GT for studying hand-based actions and object interactions with two hands. It includes multi-view RGB-D images with action labels for 36 classes constructed from verb and object labels, 3D poses for both hands giving $j = 2 \times 21$, 6D poses and meshes for manipulated objects, GT camera poses, and scene point clouds. The actions were performed by four people. The training, validation and test subsets for the action recognition and hand pose estimation tasks are provided. In the action recognition subset there are 569 clips for training, 122 for validation and 242 for testing. 

\subsubsection{\textit{\textbf{FPHA}}} This dataset contains egocentric RGB-D recordings of actions performed with one hand by a user wearing a camera. The recordings have been captured by six different people. Besides RGB-D, they include 3D hand keypoint
 annotations of a single hand with 21 keypoints giving $j = 21$, 26 object labels and 45 action classes containing a verb and an object. In the study, we follow the data split provided by the dataset authors for 600 training actions and 575 actions for evaluation.

\subsection{Metrics}

To evaluate 2D hand pose estimation task we follow standard metrics used in other studies about 2D hand pose \cite{zhang2020mediapipe, mmpose2020, santavas2020attention}. These metrics in contrast to 3D pose metrics provide information in image space. Mean End-Point Error (EPE) describes the Euclidean distance in pixels between the predicted and the GT hand keypoint, the Percentage of Correct Keypoints (PCK) answers how many predicted points have an EPE under a given threshold, normalised by the width of the hand bounding box, and the Area Under the Curve (AUC) gives PCK values over thresholds from 0 to 1. The results of action recognition are reported in terms of classification accuracy following studies \cite{cho2023transformer, Kwon_2021_ICCV, wen2023hierarchical}. For each video representing an action in a test set, a predicted action label is compared to the GT label.

\subsection{Hand Pose Evaluation}

\subsubsection{\textbf{Experiment Setup}}

\textit{EffHandNet} is trained on \textit{FreiHAND} dataset with 80/10/10 proportions for training, validation and evaluation. Optimisation is done using Stochastic Gradient Descent (SGD) and Intersection over Union (IoU) loss function with a weight decay of 10\raise0.5ex\hbox{-5} and a momentum of 0.9. After 800 epochs, the momentum is reduced to 0 and the learning rate is reduced by two. \textit{EffHandEgoNet} is trained and evaluated on \textit{H2O} and \textit{FPHA} datasets, using a similar approach, the momentum and learning rate are reduced after 60 epochs and the loss combines weighted IoU (pose) and Cross-Entropy (handness) losses of each hand. For both processes, the data is augmented with random cropping, horizontal flipping, vertical flipping, resizing, rotating and blurring following \cite{buslaev2020albumentations}. Model weights are saved for the smallest EPE in the validation subset.

\subsubsection{\textbf{Results of EffHandNet in Single-Hand Task}}

\begin{table}[t]
\caption{
Results of 2D single-hand models on \textit{FreiHAND dataset}. Referenced results are reported by the authors of the methods, while unreferenced results are computed by us using open-source implementations.}
\label{freihand_test_results}
\vspace{-0.25cm}
\begin{center}
\begin{tabular}{|l||c|c|c|c|}
\hline
Method            &Year  & PCK0.2$\uparrow$   & EPE$\downarrow$   & AUC$\uparrow$ \\
\hline
\multicolumn{5}{|c|} {\textit{test} subset from random data split 80/10/10} \\

\hline
PoseResNet50 \cite{mmpose2020}      &2020       &  99.20\%          &   3.27           &  86.8     \\
\hline

MediaPipe &2020 &71.77\%  &7.45 &79.7\\
\hline
Santavas et al. \cite{santavas2020attention}    &2020       &    -        &   4.00           &   87.0 \\        
\hline
EffHandNet  &2024  & 98.70\%    & {2.24}         & {92.1}       \\
\hline
\textbf{EffHandNet+P} &2024   & \textbf{99.32\%}    & \textbf{1.59}         & \textbf{93.5} \\
\hline
\multicolumn{5}{|c|} {\textit{final test} subset} \\
\hline
MediPipe &2020 & 81.73\% & 5.29 & 83.9 \\
\hline
PoseResNet50 &2020 & 87.48\% & 4.32 & 86.0 \\
\hline
EffHandNet   &2024           & 88.76\%           &   4.19           &  86.5               \\      
\hline
\textbf{EffHandNet+P} &2024 & \textbf{91.08\%} &  \textbf{3.67} &  \textbf{87.9} \\ 
\hline

\end{tabular}
\end{center}
\vspace{-0.4cm}
\end{table}

Table \ref{freihand_test_results} presents the results of our method compared to existing approaches for single-hand pose estimation. We follow a random dataset split in proportions of 80/10/10 following other authors \cite{mmpose2020,santavas2020attention}. We provide mean results of 10 runs to reduce the randomness factor. \textit{EffHandNet} performs comparable to other works on \textit{FreiHAND} dataset in the 2D domain with an accuracy of 98.79\% in PCK0.2, EPE equal 1.97 and AUC 92.7\%. Other studies \cite{mmpose2020,santavas2020attention} do not declare results on 2D hand keypoint prediction in a \textit{final test} subset of \textit{FreiHAND}. Our method in \textit{final test} results in better performance in all three metrics. The \textit{final test} subset with real-world background images shows that some of the predictions are outside the hand region. The reason for this is the absence of real-world background in the training samples. This is confirmed by improved results with central cropping of the images with fixed value referenced as \textit{EffHandNet+P}.

\subsubsection{\textbf{Results of 2D Egocentric Hand Pose Models}}

\begin{table}[t]
\caption{Results for 2D hand pose estimation in egocentric \textit{H2O Dataset}. The table includes hand detection accuracy, hand pose estimation PCK0.2, EPE and AUC metrics in pixels for an image size of 1280x720. Results are calculated using open-source implementations and authors' model weights.}

\begin{center}
\begin{tabular}{|l||c|c|c|c|c|}
\hline
Method:          & Year & Acc.$\uparrow$ &PCK0.2$\uparrow$ & EPE$\downarrow$ &AUC$\uparrow$\\ 
\hline
\multicolumn{6}{|c|} {\textit{H2O Dataset}} \\
\hline
PoseResNet50 \cite{mmpose2020}  & 2020 & 99.47  &74.42\%       & 26.69   &81.4\\
\hline
MediaPipe \cite{zhang2020mediapipe}& 2020 & 96.93      &86.22\% & 21.22 & 85.1\\
\hline
HTT \cite{wen2023hierarchical} & 2023   & -  & 84.75      & 19.94 & 84.8 \\
\hline

H2OTR \cite{cho2023transformer}  & 2023    & -  & 95.55      & 12.46 & 89.4 \\
\hline

EffHandNet  & 2024 & 99.47  &76.27\%      & 22.52 &82.0 \\

\hline
\textbf{EffHandEgoNet} & 2024  & \textbf{99.91}   & \textbf{97.38\%}      & \textbf{9.80} & \textbf{90.7}  \\
\hline
\multicolumn{6}{|c|} {\textit{FPHA Dataset}} \\
\hline
H2OTR \cite{cho2023transformer}  & 2023    & -  & 94.67      & 17.50 & 89.3 \\
\hline
HTT \cite{wen2023hierarchical} & 2023   & -  &92.07      & 18.07 & 88.7 \\
\hline
\textbf{Ours}  & 2024   & -  & \textbf{96.37}      & \textbf{15.20} & \textbf{88.5} \\
\hline

\end{tabular}
\end{center}
\label{tab:egocentrichandposeresults}
\vspace{-0.4cm}
\end{table}

The evaluation of the hand pose models in the egocentric perspective is performed in the test subsets of the \textit{H2O} and \textit{FPHA} datasets. The results are presented in Table \ref{tab:egocentrichandposeresults}, including hand detection accuracy, PCK.02, EPE and AUC. On \textit{H2O}, the best results are obtained by our \textit{EffHandEgoNet}, which models both hands with a PCK02 of 97.38\%, an EPE of 9.80 and an AUC of 90.7\% compared to other approaches. The performance of non-egocentric methods such as \textit{MediaPipe}, \textit{PoseResNet50} and \textit{EffHandNet} drops significantly in this scenario due to the challenging self-occluded hand frames where pose estimation is more complex. This is also visible at the hand detection stage, where \textit{MediaPipe} performs the worst, mistaking the left hand for the right [supplement]. 
In addition, \textit{EffHandEgoNet} is tested on \textit{FPHA Dataset} with state-of-the-art methods HTT \cite{wen2023hierarchical} and H2OTR \cite{cho2023transformer}, which are also outperformed. Hand detection is omitted in this part, as no labels are provided. In both datasets, H2OTR \cite{cho2023transformer} and HTT \cite{wen2023hierarchical} are transformed to image space as the original papers only provide 3D metrics.

\subsection {Action Recognition Evaluation}

\subsubsection{\textbf{Experiment Setup}}
\label{Experiment_Setup_Action}

To mitigate overfitting between the validation and training subsets, our training strategy incorporates various augmentations applied to the sequence vectors of keypoints, denoted as $\mathit V_{seq}$. 
In the process of training our method in \textit{H2O Dataset}, beneficial augmentations include horizontal and vertical flipping, random rotations, and random cropping, following Buslaev et al. \cite{buslaev2020albumentations}. Moreover, we implement an effective augmentation strategy by randomly masking either the hands or object positions.  This involves randomly setting the corresponding values of the hand or object in frame $\mathit f_n$ to zero. The models are trained with keypoints predicted from the previous stage. For \textit{FPHA Dataset} the input vector is reduced by the object bounding box and second hand to shape $\mathit f_n \in \mathbb{R}^{43}$ due to this dataset constraints. For both datasets, input sequence frames are randomly sub-sampled during training and uniformly sub-sampled for validation and testing. The models are trained with a batch size $b_s = 64$, AdamW optimiser, cross-entropy loss function, and a learning rate $l_r = 0.001$ reduced by a factor 0.5 after 900 epochs every 200 epochs for \textit{H2O Dataset}, where in \textit{FPHA Dataset} it is done after 100 and 1000 epochs. Hyperparameters and augmentations are selected based on the best-performing set in the validation subset. Every run is repeated five times to reduce the effect of random initialisation of the network and mean results with standard deviations are reported.

\subsubsection{\textbf{Results}}

\begin{table}[t]
\centering
\caption{Results in accuracy of action recognition methods on \textit{H2O} and \textit{FPHA} datasets.
Inputs of methods are: \textit{Img} stands for semantic features extracted from an image using CNN network, \textit{Hand P.} and \textit{Obj P.} stand for pose information type for hands and objects, and \textit{Obj L.} stands for object label. Results are from referenced papers.}
\label{tab:results}
\begin{tabular}{|l||c|c|c|c|c|c|}
\hline

\multicolumn{7}{|c|} {\textit{H2O Dataset}} \\
\hline

Method:   & Year & Img & H. P. & Obj P.  & Obj L.  & Acc.$\uparrow$ \\
\hline
C2D \cite{wang2018non} & 2018 & \checkmark     & \ding{55}  & \ding{55} & \ding{55}      & 70.66  \\
\hline
I3D \cite{carreira2017quo} & 2017 & \checkmark & \ding{55} &\ding{55}  & \ding{55}           & 75.21   \\
\hline
SlowFast \cite{feichtenhofer2019slowfast} & 2019 & \checkmark & \ding{55} &\ding{55}  & \ding{55}     & 77.69 \\
\hline
H+O \cite{tekin2019h+}  & 2019   &  \ding{55} & 3D  & 6D & \checkmark      & 68.88  \\
\hline
ST-GCN \cite{yan2018spatial} & 2018 &   \ding{55} & 3D & 6D & \checkmark        & 73.86\\
\hline
TA-GCN \cite{Kwon_2021_ICCV} & 2021 & \ding{55} & 3D &6D & \checkmark        & 79.25\\
\hline

HTT \cite{wen2023hierarchical}  & 2023 & \checkmark & 3D & \ding{55}   & \checkmark   & 86.36\\ 
\hline
H2OTR \cite{cho2023transformer} & 2023 & \ding{55} & 3D & 6D   & \checkmark   & 90.90\\ 
\hline

\textbf{Ours} & 2024 & \ding{55}  & \textbf{2D} & \textbf{2D} & \checkmark & \textbf{91.32} \\
\hline

\multicolumn{7}{|c|} {\textit{FPHA Dataset}} \\
\hline

Method:  & Year  & Img & H. P. & Obj P. & Obj L.  & Acc.$\uparrow$ \\
\hline
FPHA \cite{garcia2018first} & 2018 & \ding{55}  & 3D  & - & \checkmark          & 78.73\\
\hline
H+O \cite{tekin2019h+} & 2019 & \ding{55}  & 3D  & -& \checkmark & 82.43 \\
\hline
Coll. \cite{yang2020collaborative} & 2020 & \checkmark  & 3D  & -& \checkmark & 85.22 \\
\hline
HTT \cite{wen2023hierarchical} &2023  & \checkmark & 3D    & -& \checkmark          & 94.09\\ 
\hline
VPA \cite{sabater2021domain} & 2021 & \ding{55}  & 3D   & -& \checkmark & \textbf{95.93}\\
\hline
\textbf{Ours} &2024  & \ding{55} & \textbf{2D}  & - & \checkmark            & 94.43          \\
\hline

\end{tabular}
\vspace{-0.4cm}
\end{table}

We test our approach with the best performing hand pose estimation method \textit{EffHandEgoNet} and object detection implementation in \textit{H2O Dataset}. Fine-tuning of the \textit{YOLOv7} model results in 0.995 mAP@0.5, making the object data valid for higher-level inference. The proposed method results in 89.17\%$\pm$1.56, with the best model reaching an accuracy of 91.32\%, outperforming all state-of-the-art methods. 

Nevertheless, \textit{H2O Dataset} only consider scenarios of users performing actions with both hands. To prove the applicability of the proposed method in single-hand tasks, we assess our model in \textit{FPHA Dataset}, which contains actions with a single hand. Since there is no provided GT object pose for all actions, we add to the \textit{EffHandEgoNet} hand pose estimation model an additional network that predicts a class of manipulated objects. Our approach results in a mean accuracy of $93.66\%\pm0.49$ with the best result of 94.43\%, which positions our method alongside state-of-the-art methods, confirming a robust performance also in single-hand actions. 

Table \ref{tab:results} presents a comparison of state-of-the-art action recognition methods and their results in the \textit{H2O Dataset} and \textit{FPHA Dataset} reported in original studies. Following other works we report our best results. To facilitate a fair comparison, the table provides information about the input types of action recognition modules, where \textit{Img} stands for semantic features extracted from an image using CNN network, \textit{Hand P.} and \textit{Obj P.} stand for pose information type for hands and objects, and \textit{Obj L.} stands for object label. In the case of \textit{FPHA Dataset}, the comparison is done on the whole test subset where object information is not available. Due to this, some studies implemented in \textit{H2O Dataset} are omitted as they report results only on a subset with object information, making a comparison unfair. Our approach uses the least amount of information compared to other methods and still achieves the best result in \textit{H2O Dataset} and competitive accuracy in \textit{FPHA Dataset}.

\subsection{Inference Time per Action}

We measure the inference time of our method by calculating the prediction time of a single action in the \textit{H2O Dataset}. 
The study is conducted on the \textit{H2O Dataset} specifically, as it presents a more complex scenario involving interactions with objects by both hands. This differs from the \textit{FPHA} dataset, where the reported results in Table \ref{tab:results} do not incorporate object pose information. The evaluation is performed by averaging inference times over 1000 trials on NVIDIA GeForce RTX 3090 GPU, ensuring robustness and reliability. Our method is compared with HTT \cite{wen2023hierarchical} and H2OTR \cite{cho2023transformer} methods based on 3D hand pose input, as they are the only open-source implementations that allow such a comparison on the \textit{H2O Dataset} at the time of this study.

\begin{figure}[t]
  \centering
  \includegraphics[width=1\linewidth]{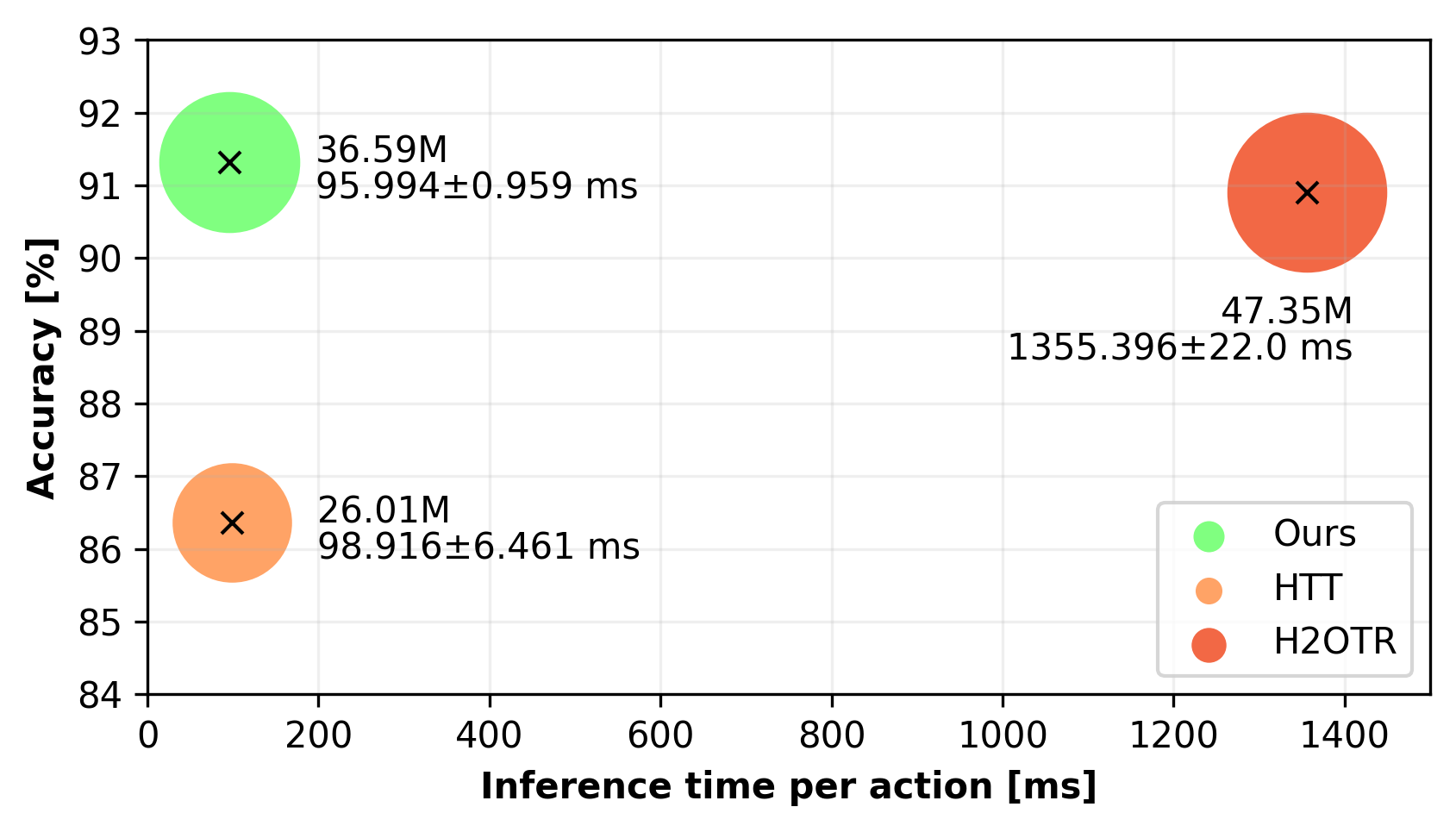}
  \caption{Inference time and accuracy per single action of state-of-the-art methods on \textit{H2O Dataset}. Our method predicts the fastest with the highest accuracy.}
  \vspace{-0.25cm}
  \label{fig:inference_time}
\end{figure}

The results demonstrate the improvement of our method over HTT \cite{wen2023hierarchical} and H2OTR \cite{cho2023transformer} in terms of inference speed and accuracy. We achieve an accuracy of 91.32\% with an inference time of $95.994\pm0.959$ ms. In contrast, HTT \cite{wen2023hierarchical} needs $98.916\pm6.461$ to reach 86.36\% and H2OTR \cite{cho2023transformer} with 90.90\% is about 14 times slower and performs the inference in $1355.396\pm22$ ms. The results are shown in Fig. \ref{fig:inference_time}, where each method is visualised as a circle whose size depicts the number of its parameters.

\subsection{Ablation Study}

Various ablations regarding input data types are performed to expand the evaluation of our approaches. We present the study of the action recognition performance with distinct methods for hand pose estimation and the impact of each input, i.e., left or right hand, on the final accuracy. All experiments are performed in \textit{H2O Dataset} with fixed seeds to ensure reproducibility. 

\subsubsection{\textbf{EffHandEgoNet Feature Extractor}}

To find the best-performing future extractor, we experiment with different backbones. The selection is based on network size and architecture type. The results are shown in the Table \ref{tab:ablation_backbone}. The best performance is observed with \textit{EfficientNetV2S} \cite{tan2021efficientnetv2}, but it is associated with the highest inference time. The faster alternative are \textit{ConvNext Tiny} \cite{liu2022convnet} followed by \textit{ResNet50} \cite{he2016deep} or \textit{MobileNetV3 Large} \cite{howard2019searching}. The worst performance is observed for \textit{SwinV2T Tiny}. \cite{liu2022swin}, which overfits the test data. All inference times are measured on the RTX 3090 GPU.

\begin{table}[t]
\caption{Results for 2D hand pose estimation in egocentric \textit{H2O dataset}. }
\label{tab:ablation_backbone}
\begin{center}
\begin{tabular}{|l||c|c|c|c|c|}

\hline
Model:          & Param. & Inf.[ms]  &PCK0.2$\uparrow$ & EPE$\downarrow$ &AUC$\uparrow$\\
\hline
SwinV2T \cite{liu2022swin} & 31.4M &16.02  & 85.76\% & 18.87 & 0.851\\
\hline
MobileNetV3 \cite{howard2019searching} & 7.3M &7.59 & 92.95\% & 13.21 & 0.884\\
\hline
ResNet50 \cite{he2016deep}  & 30.3M & 6.96    & 95.46\% & 11.66 & 0.895\\
\hline
ConvNext \cite{liu2022convnet}  & 31.7M &7.17 & 95.67\% & 11.48 & 0.896\\
\hline
\textbf{Eff-NetV2S} \cite{tan2021efficientnetv2} & 25.2M &21.47   &\textbf{97.38\%}      & \textbf{9.80} &\textbf{0.907}  \\
\hline
\end{tabular}
\end{center}
\end{table}

\subsubsection{\textbf{Hand pose in frames with overlapping hands}}

\begin{figure}[t]
  \centering
  \includegraphics[width=1\linewidth]{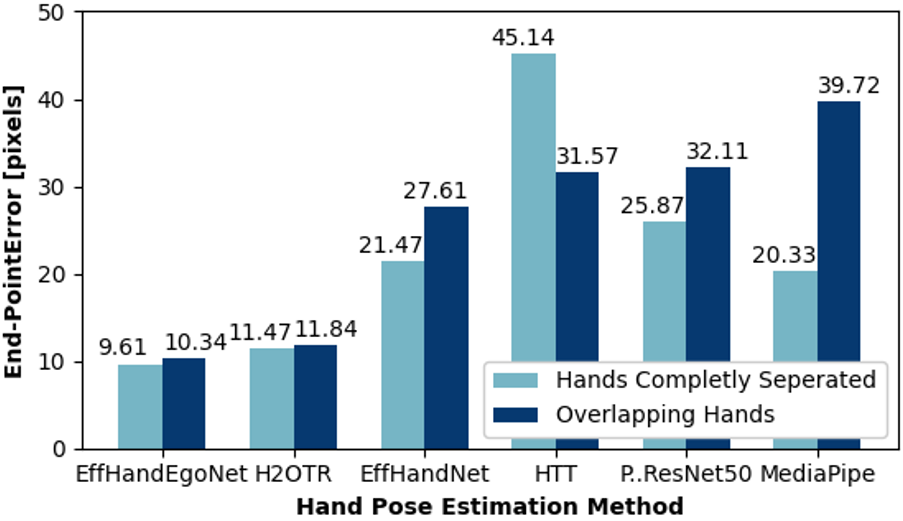}

  \caption{EPE results for different methods in edge scenarios for overlapping and fully separated hands in \textit{H2O Dataset}. }
  \label{fig:ablation_distance}
\end{figure}

To understand how different models perform with self-occlusion, we divide the test data in egocentric \textit{H2O Datasets} into two edge scenarios. The first subset consists of frames in which the hands of individuals performing actions are separated by a distance of 30\% of the image width, and the second includes hands that overlap by at least 10\% of the image width. The results presented in Fig. \ref{fig:ablation_distance} of the EPE for each method in these scenarios show the superior performance of our \textit{EffHandEgoNet}. Not only its performance is the best, but the gap between scenarios is marginal, proving the robustness of our method to self-occlusions.

\subsubsection{\textbf{Action Recognition and Hand Pose Models}}

\begin{figure}[t]
  \centering
\includegraphics[width=\linewidth]{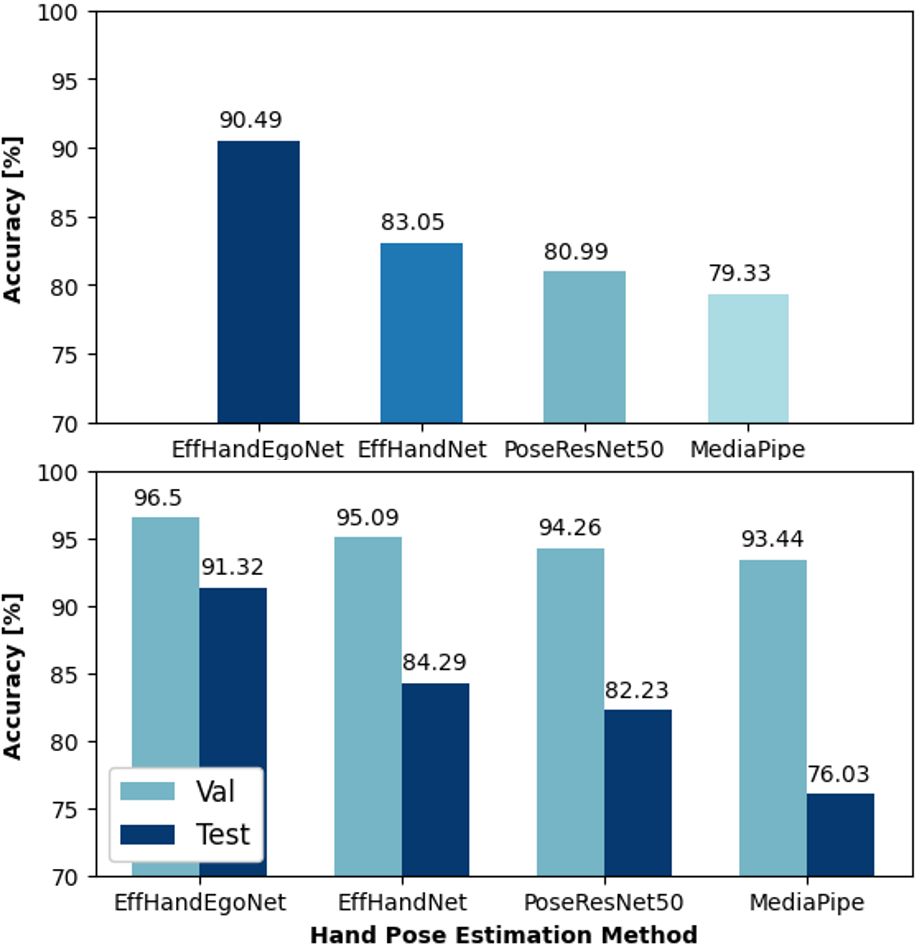}
  \caption{Accuracy of action recognition depending on hand pose methods in \textit{H2O Dataset}. On the top model trained with GT pose and tested with the estimated pose, on the bottom trained and tested with the estimated pose.}
  \label{ablations_handmodel}
\end{figure}

The challenge of estimating hand poses from an egocentric perspective, where the movement involves the user performing actions, is notable due to occlusions from both the hands and the manipulated objects. This complexity leads to a decrease in performance when moving from a non-egocentric to an egocentric viewpoint, as evidenced by the different results presented in Table \ref{freihand_test_results} and Table \ref{tab:egocentrichandposeresults}. We investigate the impact of hand pose estimation methods on action recognition accuracy in two ways. 

In our first experiment, the action recognition model is trained using GT poses following the same strategy outlined in Section \ref{Experiment_Setup_Action}, which results in a validation accuracy of 96.72\% and 92.97\% for the test. During the testing phase, different hand pose estimation methods are used to evaluate how their differing levels of pose estimation precision influence the final results. The results for action recognition show a correlation with the accuracy of hand pose estimation, ranging from a low of 79.33\% for the \textit{MediaPipe} network to a high of 90.49\% for the \textit{EffHandEgoNet} network. The results of each method are shown in the upper panel of Fig.\ref{ablations_handmodel}.

In the second one, the action recognition network is trained from scratch for each hand pose estimation method using the predicted poses. While close performance is observed in the validation subsets, the test set shows a similar correlation between hand pose methods as observed in our first experiment (lower panel in Fig.\ref{ablations_handmodel}).

Performance in action recognition is shown to be directly related to hand pose estimation precision, which highlights the importance of accurate hand pose input, particularly in egocentric scenarios where actions involve interactions between both hands manipulating objects.

\subsubsection{\textbf{Different Network Inputs for Action Recognition}}

Experiments with each network input highlight their respective importance for action recognition accuracy, as shown in Table \ref{tab:ablations2} alongside a comparison to a complete network. To streamline our analysis and avoid the influence of estimation algorithms for each network part, we perform this experiment using GT information, involving the ablation of various network inputs. \textit{GTPose III} does not incorporate an object bounding box, reducing the performance to 76.03\%, which emphasises the importance of including object pose information. \textit{GTPose IV} contains only left hand pose and object information and results in 73.14\%. \textit{GTPose V} uses the right hand with object information and results in 79.33\%. The disparity in accuracy for hand type underlines that one of the hands plays a more influential function in human actions, but information regarding both poses is critical for successful performance. 

\begin{table}[!t]
\centering
\caption{Results of our ablation study depending on inputs: \textit{left hand pose}, \textit{right hand pose} and \textit{object pose}.}
\label{tab:ablations2}
\begin{tabular}{|l||c|c|c|c|}

\hline

Method:    & Left Hand & Right Hand &Obj Pose   & Acc. {[}\%{]} $\uparrow$ \\ 
\hline
GT &  \checkmark & \checkmark & \checkmark  & 92.97 \\
\hline
GTPose III &\checkmark& \checkmark & \ding{55}                 & 76.03 \\
\hline
GTPose IV &\checkmark& \ding{55} & \checkmark               &   73.14 \\
\hline
GTPose V &\ding{55} & \checkmark & \checkmark             & 79.33 \\
\hline
\end{tabular}
\vspace{-0.4cm}
\end{table}

\section{Conclusion}
\label{conclusion}

Under this study, two novel 2D hand pose estimation models were developed to target the challenges of the egocentric perspective. 
The top-down approach, \textit{EffHandNet}, resulted in the improvement of all three used metrics in the single-hand \textit{FreiHAND} dataset and state-of-the-art performance in the egocentric \textit{H2O Dataset}. The bottom-up \textit{EffHandEgoNet} model improved state-of-the-art performance in every metric on egocentric \textit{H2O} and \textit{FPHA} datasets, accurately estimating the most challenging scenarios, including overlapping hands and occlusions. 
Further, estimated hand poses alongside object detection were used in a novel egocentric action recognition model, where each frame is described by a vector containing the hand pose and the object bounding box. The sequence of vectors is processed by a transformer-based neural network. Evaluation of our model on \textit{FPHA Dataset} showed competitive performance with 94.43\% accuracy, confirming accurate performance for actions performed with a single hand. Results obtained on \textit{H2O Dataset}, where two hands are involved in the action, resulted in 91.32\% accuracy, outperforming state-of-the-art and achieving a faster inference time.

Additional experiments and ablations demonstrated the impact of hand pose estimation methods. Approaches originally designed for a non-egocentric perspective, despite state-of-the-art performance in single-hand datasets, showed a decrease in performance on egocentric pose estimation and action recognition tasks compared to our egocentric approach \textit{EffHandEgoNet}. We showed that accurate pose description is essential for correct action understanding and that to achieve the best performance, it is necessary to use a method capable of correctly modelling interactions between hands and manipulated objects like \textit{EffHandEgoNet}. Our study demonstrates that in certain scenarios, i.e. reducing model complexity to reduce inference speed, 2D pose information is a promising alternative to estimated 3D pose for egocentric action recognition, as it competes strongly with non-pose and 3D hand pose-based methods.


\section{Acknowledgements}

This work was supported by VisuAAL ITN H2020 (grant agreement No. 861091) and by Vienna Science and Technology Fund (grant agreement No. ICT20-055).

{\small
\bibliographystyle{ieee}
\bibliography{egbib}
}

\end{document}